%% file: main_wacv.tex
\pdfoutput=1
\documentclass[10pt,twocolumn,letterpaper]{article}

\usepackage[applications]{wacv}

\input{preamble}

\input{tables/result_numbers}
\input{tables/data_numbers}
\input{tables/merge_numbers}

\usepackage[T1,OT1]{fontenc}
\DeclareTextSymbolDefault{\textquotedbl}{T1}
\usepackage{listings}
\definecolor{wacvblue}{rgb}{0.21,0.49,0.74}
\usepackage[pagebackref,breaklinks,colorlinks,allcolors=wacvblue]{hyperref}

\def\wacvPaperID{31}
\def\confName{WACV}
\def\confYear{2027}

\title{Generalist vs.\ Specialist Vision Models for Building Attributes:\\
A Cross-Region Evaluation with Time-Checked Labels}

\author{%
Seyedeh Mobina Noorani$^{1,2}$\qquad Karla Salda\~{n}a Ochoa$^{1}$\\[4pt]
{\normalsize $^{1}$College of Design, Construction and Planning, University of Florida, Gainesville, FL 32611, USA}\\
{\normalsize $^{2}$Department of Electrical and Computer Engineering, University of Florida, Gainesville, FL 32611, USA}%
}

\begin{document}
\maketitle
\AddToShipoutPicture*{\AtTextLowerLeft{\put(0,\LenToUnit{-0.72in}){\parbox[b]{\textwidth}{\scriptsize\centering
Accepted at the GeoCV Workshop, 2027 IEEE/CVF Winter Conference on Applications of Computer Vision Workshops (WACVW).\\
\copyright\ 2027 IEEE. Personal use of this material is permitted. Permission from IEEE must be obtained for all other
uses, in any current or future media, including reprinting/republishing this material for advertising or promotional
purposes, creating new collective works, for resale or redistribution to servers or lists, or reuse of any copyrighted
component of this work in other works.}}}}
\input{sec/0_abstract}
\input{sec/1_intro}

\input{sec/2_related}

\input{sec/3_data}

\input{sec/4_methods}

\input{sec/5_results}

\input{sec/6_discussion}
{
    \small
    \bibliographystyle{ieeenat_fullname}
    \bibliography{refs_wacv}
}

\setcounter{section}{0}
\setcounter{table}{0}
\setcounter{figure}{0}
\renewcommand{\thesection}{S\arabic{section}}
\renewcommand{\thetable}{S\arabic{table}}
\renewcommand{\thefigure}{S\arabic{figure}}
\renewcommand{\theHsection}{S\arabic{section}}
\renewcommand{\theHtable}{S\arabic{table}}
\renewcommand{\theHfigure}{S\arabic{figure}}
\twocolumn[{\centering\Large\bfseries Supplementary Material\par\vspace{1.5em}}]
\input{sec/7_supplement}

\end{document}

%% file: preamble.tex
\usepackage{graphicx}
\usepackage{amsmath,amssymb}
\usepackage{booktabs}
\usepackage{multirow}
\usepackage{enumitem}  %
\usepackage{xcolor}
\newcommand{\TODO}[1]{\textbf{\color{red}[TODO: #1]}}
\newcommand{\NUM}[1]{{\color{blue}#1}}
\renewcommand{\TODO}[1]{}
\renewcommand{\NUM}[1]{#1}

%% file: tables/result_numbers.tex
\newcommand{\ResultUkFloorsQwen}{80.7}

\newcommand{\ResultUkFloorsOfmodel}{89.4}

\newcommand{\ResultUkLowFloorsGemmaMoe}{98.2}

\newcommand{\ResultLondonWallClaude}{92.7}

\newcommand{\ResultLondonWallBmat}{87.8}

\newcommand{\ResultLondonWallCount}{160}
\newcommand{\ResultScotlandWallClaude}{93.1}

\newcommand{\ResultScotlandWallBmat}{93.8}

\newcommand{\ResultScotlandWallCount}{107}

\newcommand{\ResultOfFloorsGemmaMoe}{71.4}

\newcommand{\ResultOfFloorsCount}{102}
\newcommand{\ResultOfYearClaude}{48.6}

\newcommand{\ResultOfWallCount}{135}

\newcommand{\ResultOfShapeCount}{128}

\newcommand{\ResultOfThreeShapeGemini}{78.1}

\newcommand{\ResultOfThreeShapeBrails}{64.1}
\newcommand{\ResultOfThreeShapeCount}{81}

\newcommand{\ResultOfRoofRoofnet}{48.2}
\newcommand{\ResultOfRoofCount}{81}

\newcommand{\ResultFloridaFloorsQwen}{89.5}

\newcommand{\ResultFloridaFloorsBrails}{61.6}
\newcommand{\ResultFloridaFloorsOfmodel}{90.8}
\newcommand{\ResultFloridaFloorsCount}{134}
\newcommand{\ResultFloridaYearClaude}{51.9}

\newcommand{\ResultFloridaYearQwen}{28.7}

\newcommand{\ResultFloridaYearBrails}{21.0}

\newcommand{\ResultFloridaYearCount}{169}
\newcommand{\ResultFloridaEraClaude}{94.7}
\newcommand{\ResultFloridaEraLo}{85.9}
\newcommand{\ResultFloridaEraHi}{100.0}

\newcommand{\ResultFloridaEraCount}{169}
\newcommand{\ResultFloridaWallClaude}{76.9}

\newcommand{\ResultFloridaWallBmat}{63.8}

\newcommand{\ResultFloridaWallCount}{133}

\newcommand{\ResultFloridaRoofRoofnet}{15.3}
\newcommand{\ResultFloridaRoofCount}{139}
\newcommand{\ResultLeadingRowsClaude}{10}
\newcommand{\ResultLeadingRowsGemini}{2}
\newcommand{\ResultLeadingRowsGemmaMoe}{2}
\newcommand{\ResultLeadingRowsBmat}{1}
\newcommand{\ResultPhotoFloorsCount}{131}
\newcommand{\ResultPhotoFloorsGain}{9.6}
\newcommand{\ResultPhotoFloorsLo}{1.1}
\newcommand{\ResultPhotoFloorsHi}{19.5}
\newcommand{\ResultPhotoWallCount}{116}
\newcommand{\ResultPhotoWallGain}{15.3}
\newcommand{\ResultPhotoWallLo}{-0.6}
\newcommand{\ResultPhotoWallHi}{35.6}
\newcommand{\ResultPhotoRoofCount}{74}
\newcommand{\ResultPhotoRoofGain}{15.6}
\newcommand{\ResultPhotoRoofLo}{-0.2}
\newcommand{\ResultPhotoRoofHi}{25.5}

\newcommand{\ResultFloorScreenDropClaude}{5.9}
\newcommand{\ResultFloorScreenDropGemini}{7.1}
\newcommand{\ResultFloorScreenDropQwen}{6.7}
\newcommand{\ResultFloridaNewEraCount}{10}
\newcommand{\ResultUkVote}{89.2}
\newcommand{\ResultUkVoteClaude}{92.6}
\newcommand{\ResultOfVote}{66.7}
\newcommand{\ResultOfVoteClaude}{67.5}
\newcommand{\ResultVoteGapMin}{0.8}
\newcommand{\ResultVoteGapMax}{20.1}
\newcommand{\ResultAgreeFlagMin}{3.0}
\newcommand{\ResultAgreeFlagMax}{45.1}

%% file: tables/data_numbers.tex
\newcommand{\DataUkTestFloors}{121}
\newcommand{\DataUkTestWalls}{267}
\newcommand{\DataUkPracticeFloors}{110}
\newcommand{\DataUkPracticeWalls}{251}
\newcommand{\DataUkStreetQuestions}{385}
\newcommand{\DataOfTestPool}{382}
\newcommand{\DataOfPracticePool}{213}
\newcommand{\DataOfRoofQuestions}{150}
\newcommand{\DataOfUnknownCity}{65}
\newcommand{\DataFloridaPool}{348}
\newcommand{\DataFloridaPractice}{175}
\newcommand{\DataFloridaTest}{173}
\newcommand{\DataLidarComparedFloors}{279}
\newcommand{\DataLidarAgreeFloors}{260}
\newcommand{\DataLidarComparedRoof}{258}
\newcommand{\DataLidarAgreeRoof}{256}
\newcommand{\DataCmpTest}{114}
\newcommand{\DataCmpPractice}{114}

%% file: tables/merge_numbers.tex
\newcommand{\MergeMainRows}{15}

\newcommand{\MergeSpecialistRows}{22}
\newcommand{\MergeSpecialistBehind}{15}

\newcommand{\CtrlOfRoofClaudeN}{81}
\newcommand{\CtrlOfRoofClaudeStreet}{41.7}
\newcommand{\CtrlOfRoofClaudeControl}{55.7}
\newcommand{\CtrlOfRoofClaudeBoth}{80.9}
\newcommand{\CtrlOfRoofClaudeAerialGain}{25.5}
\newcommand{\CtrlOfRoofClaudeAerialLo}{13.9}
\newcommand{\CtrlOfRoofClaudeAerialHi}{37.0}
\newcommand{\CtrlOfRoofClaudeQuestionGain}{14.0}
\newcommand{\CtrlOfRoofClaudeQuestionLo}{4.4}
\newcommand{\CtrlOfRoofClaudeQuestionHi}{24.0}

\newcommand{\CtrlOfRoofGeminiAerialGain}{13.2}
\newcommand{\CtrlOfRoofGeminiAerialLo}{-0.5}
\newcommand{\CtrlOfRoofGeminiAerialHi}{27.1}
\newcommand{\CtrlOfRoofGeminiQuestionGain}{19.1}

\newcommand{\CtrlOfShapeClaudeAerialGain}{-5.2}
\newcommand{\CtrlOfShapeClaudeAerialLo}{-14.1}
\newcommand{\CtrlOfShapeClaudeAerialHi}{3.9}
\newcommand{\CtrlOfShapeClaudeQuestionGain}{8.3}
\newcommand{\CtrlOfShapeClaudeQuestionLo}{2.6}
\newcommand{\CtrlOfShapeClaudeQuestionHi}{14.3}

\newcommand{\CtrlFlRoofClaudeN}{139}

\newcommand{\CtrlFlRoofClaudeAerialGain}{4.5}
\newcommand{\CtrlFlRoofClaudeAerialLo}{-1.8}
\newcommand{\CtrlFlRoofClaudeAerialHi}{11.8}

%% file: sec/0_abstract.tex
\begin{abstract}
Evaluating building vulnerability to climate-related hazards requires accurate physical attributes, often missing or outdated in property records. Street-level, aerial, and satellite imagery offer alternative sources, while AI vision models can automate their extraction. However, their reliability across regions and building characteristics remains uncertain. This study evaluates ten general-purpose vision-language models and four specialized models for extracting floor count, construction period, wall and roof materials, and roof shape. We use existing datasets from London, Scotland, Amsterdam, Berlin, Helsinki, San Francisco, Washington, D.C., Houston, and Manila, alongside a new Gainesville, Florida dataset whose county-record labels are screened with LiDAR estimates and building permits.

Results show that no model clearly outperforms Claude Sonnet 5.5, while specialized models remain competitive mainly on selected tasks. Reference-data screening substantially influences measured performance, with floor-classification scores decreasing by six to seven points when LiDAR agreement screening is removed. Aerial imagery improves roof-material classification in some settings, while promptable segmentation models remain competitive for facade-object counting. Model disagreement may also help prioritize manual review. Overall, reliable building-attribute extraction depends on model selection, imagery, task formulation, and reference quality. These findings provide practical guidance for scalable AI-based building characterization supporting preliminary climate-hazard vulnerability assessments.
\end{abstract}

%% file: sec/1_intro.tex
\section{Introduction}
\label{sec:intro}

Climate-related hazards, including flooding, hurricanes, and extreme weather events, increasingly threaten buildings and urban infrastructure. Assessing the potential vulnerability of the built environment requires detailed information about individual buildings, including their physical characteristics, materials, construction period, and geometry; for other infrastructure, asset age is a leading predictor of condition~\cite{mohammadagha2025pipes}. Attributes such as floor count, wall and roof materials, and roof shape can inform hazard exposure and vulnerability assessments by describing characteristics that influence how buildings may respond to environmental stresses. However, these data are frequently incomplete, inconsistent~\cite{hazus,brails}, or outdated in administrative property records, limiting the scalability and reliability of building-level assessments and of the regional adaptation planning built on them~\cite{noorani2026regionalization,noorani2025agentic}. The growing availability of street-level, aerial, and satellite imagery provides opportunities to address these information gaps. Street-view photographs capture visible facade characteristics, including wall materials, openings, and architectural features, while aerial and satellite images provide complementary information about roof geometry and materials. Traditionally, extracting these attributes from imagery has required manual interpretation or task-specific computer vision models~\cite{brails,openfacades,bmat,tarver2025roofnet}, which makes large-scale application resource-intensive. Vision-language models (VLMs) offer an alternative: they extract building characteristics from natural-language prompts without task-specific training, although LLM choices can change with the framing of an otherwise identical task~\cite{davoudi2026samegame}. Despite this potential, published results use different datasets, classes and labels, so it remains unclear which model to use. Answering this requires comparing models on the same photos, both within and outside the training regions of specialists.

A second problem concerns the answer key. A county record describes the building today, while the photo may be
years old, and a single administrative record can be wrong. In our Florida sample, about half of the roof-material
records fail a photo-date check against the crowd-sourced street photos, mostly because of a later re-roofing
permit (\cref{sec:labels}). We therefore
screen the Florida labels with airborne LiDAR and with permit dates, and measure how much these screens change the
scores.

\begin{figure*}[t]
  \centering
  \includegraphics[width=\linewidth]{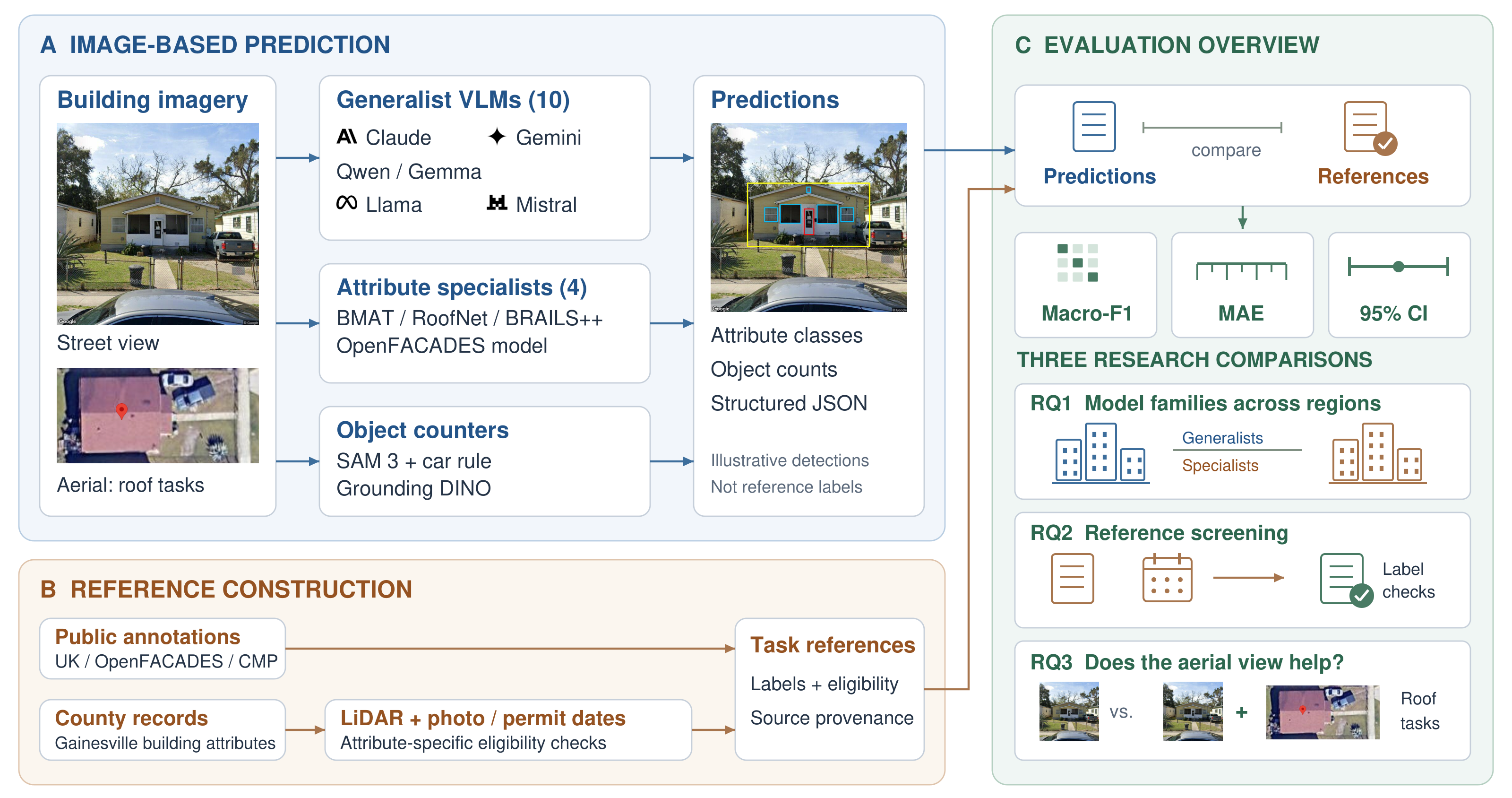}
  \caption{Study overview. (A)~Street photos, plus aerial photos for the roof tasks, go to ten zero-shot generalists,
  four released specialists and two object counters; the boxes on the prediction photo are illustrative, not
  reference labels. Counting is evaluated only on CMP facades, with SAM~3, Grounding DINO, Claude, Gemini and Qwen. (B)~References come from public labels (UK, OpenFACADES, CMP) and from Gainesville county
  records screened with LiDAR and photo and permit dates. (C)~Predictions are compared with the references by
  macro-F1, mean absolute error for counts and 95\% bootstrap intervals, organized around RQ1--RQ3.
  Example imagery \copyright~Google.}
  \label{fig:overview}
\end{figure*}

We examine five building attributes: floor count, construction year, wall material, roof shape, and roof material. Our analysis uses datasets from London and Scotland in the UK, seven cities represented in OpenFACADES, and Gainesville, Florida, USA. We also assess the counting of windows, doors, and balconies using rectified facade images. Our research questions are:
\begin{itemize}[leftmargin=*,itemsep=1pt,topsep=2pt]
  \item \textbf{RQ1, generalists and specialists.} On matched evaluation cases, using each model's supported image inputs, how do zero-shot generalists compare with released specialists inside and outside the regions and image sources represented in specialist training data? Can the generalists also count facade objects as well as promptable detectors?
  \item \textbf{RQ2, label validity.} How much do screening county labels with LiDAR and permit dates, and the
        age of the photos, change measured scores and the ranking of models?
  \item \textbf{RQ3, two views.} With the question held fixed, does adding an aerial photo to the street photo
        improve the roof attributes?
\end{itemize}
\Cref{fig:overview} summarizes the study design.

Our contributions are:
\begin{itemize}[leftmargin=*,itemsep=1pt,topsep=2pt]
  \item a cross-region evaluation of ten zero-shot generalists (two commercial, eight open) and four released specialist models on matched cases across five building attributes, plus a separate comparison of three generalists and two promptable detectors for facade counting, with documented class mappings, split-overlap checks and paired statistics;
  \item time-checked labels for \NUM{\DataFloridaPool} Florida houses, built from public property records,
        building permits and airborne LiDAR, and a measurement of how these screens and the age of the photos change
        the scores;
  \item practical guidance based on the evaluated conditions, including model selection by attribute, the benefits of aerial imagery for roof material, the role of promptable detectors in facade counting, and the potential use of model disagreement to identify cases for manual review.
\end{itemize}
Because region, image source and label source change together across our datasets, we report how performance
differs between these settings, not a single cause for each difference. We release the code, the questions, the Florida labels
and a simple tool that applies the recommended setup to new photos.

%% file: sec/2_related.tex
\section{Related work}
\label{sec:related}

\paragraph{Building attributes from images.}
Street view imagery is now an established source for creating survey data to describe the built environment~\cite{biljecki2021}. For
natural-hazard simulation, BRAILS~\cite{brails,brailspp} trains task-specific networks on US street-level and
satellite images to predict building attributes such as floors, year built and roof shape at regional scale.
Zero-shot pipelines have combined CLIP and SAM for the same attributes~\cite{zeroshot}. OpenFACADES~\cite{openfacades}
released \NUM{31,180} street-view building images
from seven cities (Amsterdam, Berlin, Helsinki, San Francisco, Washington, D.C., Houston, and Manila), with attribute labels derived from
geospatial data, and a fine-tuned vision-language model.
BMAT~\cite{bmat} fine-tuned a vision-language
model on Google and Baidu street-view images of 73 cities to map facade materials, RoofNet~\cite{tarver2025roofnet} fine-tuned RemoteCLIP to
recognize roof materials in satellite tiles, and BuildingMultiView~\cite{buildingmultiview} fine-tuned a large model on
paired satellite and street views. Existing studies differ in evaluated attributes, class definitions, image sources and reference data, making direct comparisons of their reported results difficult. We address this gap by evaluating released generalist and specialist models on matched cases within each attribute and dataset, and examining how reference screening affects measured performance.

\paragraph{General vision-language models for urban scenes.}
Benchmarks of large multimodal models on urban scenes find
that counting and object attributes remain challenging,
and that model behavior can be inconsistent across
different urban views~\cite{urbench}.
A fine-tuned Gemma 3 27B model shows strong agreement
with mean human ratings of built-environment and housing attributes~\cite{yao2026},
while recent work examines what street-view imagery adds
beyond existing urban data~\cite{tan2026}.
We add an exposure-oriented comparison of zero-shot
generalist and specialist models across regions.
For counting facade objects, Grounding DINO and SAM 3
detect or segment objects specified by text
prompts~\cite{groundingdino,sam3}.
We evaluate them without facade-specific fine-tuning
against generalist models on annotated
facades~\cite{cmp}.

\paragraph{Labels and their timing.}
Reference sources and validation procedures differ across datasets;
BMAT, for example, assesses agreement with Overture Maps building
records~\cite{bmat}. Our Florida analysis examines how permit-date
screening and agreement with a LiDAR-derived estimate change the
evaluated cohorts and measured scores (\cref{sec:labels}).

%% file: sec/3_data.tex
\section{Data and references}
\label{sec:data}

We evaluate floors, construction period, wall material, roof shape and roof material where references are available.
The sources are expert annotations, provided dataset attributes, OpenStreetMap (OSM) tags and administrative records
(\cref{tab:data}). Their reliability and coverage differ; we do not treat them as equivalent independently verified
ground truth. Windows, doors and balconies are evaluated separately on rectified facades.

\paragraph{UK.}
The building-characteristics dataset~\cite{ukchar} provides expert-labeled Street View images from London and
Scotland. We use its test folders after within-task image-hash deduplication. Supported references comprise
\NUM{\DataUkTestFloors} floor examples, all in London, and \NUM{\DataUkTestWalls} wall examples:
\NUM{\ResultLondonWallCount} in London and \NUM{\ResultScotlandWallCount} in Scotland.
Mixed and other cladding labels, and unsupported floor labels, are not scored.

\paragraph{OpenFACADES.}
The source dataset~\cite{openfacades} combines Mapillary building images and geospatial attributes across seven
cities. Our selection contains \NUM{\DataOfTestPool} test and \NUM{\DataOfPracticePool} practice buildings, one image
each, chosen with class quotas (Supp.\ Sec.~S2); it is not an official test of the released OpenFACADES model.
Its floor, construction-year and wall-material attributes come mainly from OSM tags, harmonized with other open
building data~\cite{openfacades}; they are not image annotations. Roof references come from an OSM lookup, and the
aerial inputs are city orthophotos with known building outlines.

\paragraph{Gainesville.}
The county-record sample contains \NUM{\DataFloridaPool} residential parcels, divided into
\NUM{\DataFloridaPractice} practice and \NUM{\DataFloridaTest} test parcels, stratified by construction period.
Each test parcel has Google Street View and satellite inputs, and a separate Mapillary/NAIP image pair.
Mapillary crops were selected using a Qwen visibility screen; Google images did not undergo that screen.
County attributes are administrative references for the selected building, not new image annotations.
The main Google cohorts retain \NUM{\ResultFloridaFloorsCount} floor,
\NUM{\ResultFloridaYearCount} year, \NUM{\ResultFloridaWallCount} wall and
\NUM{\ResultFloridaRoofCount} roof-material references after screening.

\paragraph{Facade counts.}
We use \NUM{\DataCmpPractice} practice and \NUM{\DataCmpTest} test images from the supplied public CMP facade
copy~\cite{cmp}. Count references are the numbers of four-connected regions of each class in its segmentation maps,
an instance-count proxy: touching objects can merge and fragmented regions can split.

\paragraph{Classes and splits.}
Floors are grouped as 1, 2, 3, 4 and 5+ (1, 2 and 3+ in Florida); construction year by decade from before 1970 to
2010 or later, plus a before/after-2002 split in Florida, a calendar-year proxy rather than code compliance. Walls
merge concrete with plaster and wood with siding; the full mappings are in Supp.\ Sec.~S2. Practice and test sets are separated by image hash (UK),
building ID (OpenFACADES), and parcel ID (Florida).
However, some Florida practice and test parcels share
source images (Supp.\ Sec.~S2). The OpenFACADES specialist is not scored on its own source dataset.

\begin{table}[t]
  \centering
  \footnotesize
  \setlength{\tabcolsep}{2pt}
  \begin{tabular}{@{}p{0.23\linewidth}p{0.19\linewidth}p{0.12\linewidth}p{0.35\linewidth}@{}}
    \toprule
    Dataset & References & Pool & Main scored cohorts \\
    \midrule
    UK & experts & \NUM{\DataUkStreetQuestions} & floors \NUM{\DataUkTestFloors}; walls \NUM{\DataUkTestWalls} \\
    OpenFACADES & attributes / OSM & \NUM{\DataOfTestPool} & floors and year \NUM{\ResultOfFloorsCount} each; walls \NUM{\ResultOfWallCount}; roof shape \NUM{\ResultOfShapeCount}; roof material \NUM{\ResultOfRoofCount} \\
    Gainesville & county, screened & \NUM{\DataFloridaTest} & floors \NUM{\ResultFloridaFloorsCount}; year \NUM{\ResultFloridaYearCount}; walls \NUM{\ResultFloridaWallCount}; roof material \NUM{\ResultFloridaRoofCount} \\
    CMP facades & map regions & \NUM{\DataCmpTest} & windows, doors, balconies \\
    \bottomrule
  \end{tabular}
  \caption{Evaluation pools and attribute-specific sample counts. UK and OpenFACADES pools count street images;
  Gainesville counts parcels with two image pairs; CMP counts facades. UK tasks overlap in three image IDs.
  Roof questions exist for \NUM{\DataOfRoofQuestions} OpenFACADES buildings, but fewer have eligible references.
  The main table specifies each comparison's exact $n$.}
  \label{tab:data}
\end{table}

\subsection{Florida reference screening}
\label{sec:labels}
The floor screen uses agreement between county stories and a cached LiDAR estimate from USGS 3DEP
point clouds~\cite{threedep}. The code converts median roof height above nearby ground to a rounded floor estimate,
using three meters per floor and a 1.2-meter offset. Roof-slope estimates below five degrees are flat, at least ten
degrees are sloped, and intermediate estimates are uncertain. Cached flight-year and permit checks determine which
LiDAR estimates can be compared with county records. Across the full parcel sample, floors agree in
\NUM{\DataLidarAgreeFloors} of \NUM{\DataLidarComparedFloors} comparable cases; flat/sloped roofs agree in
\NUM{\DataLidarAgreeRoof} of \NUM{\DataLidarComparedRoof}. These are agreement counts with geometric heuristics,
not annotation-accuracy estimates. The main floor analysis retains LiDAR-agreement cases. Although LiDAR provides a binary flat/sloped roof check, Florida roof shape is omitted from the main evaluation because too few test buildings have flat roofs.

The street-photo date screen uses the saved county permit archive~\cite{acpa}. A reference is excluded if the recorded construction year is later than the photo year or if a relevant permit is dated unambiguously after the photo's capture interval.
Uncertain within-interval cases are retained.
Cases without a usable street-photo date are also excluded.
Re-roof permits affect roof material; addition/remodel permits affect the other attributes.
Google dates have month precision and Mapillary dates year precision (a stricter variant is in Supp.\ Sec.~S2).
A permit does not establish completion or material change, and aerial acquisition dates are not established.

\begin{figure}[t]
  \centering
  \includegraphics[width=\linewidth]{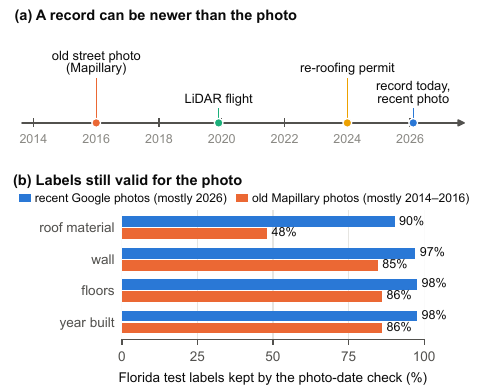}
\caption{(a) Illustration: a re-roofing permit dated after
the street photo means today's record may not describe
the photographed roof; it flags a possible mismatch,
not a proven one.
(b) Share of Florida test references retained after
photo-date screening, before the LiDAR check, for older
and more recent photos of the same houses.
Retained references are not necessarily verified
ground truth.}
  \label{fig:labels}
\end{figure}

%% file: sec/4_methods.tex
\section{Models and evaluation protocol}
\label{sec:methods}

\paragraph{Generalists and prompts.}
We use Claude Sonnet 5.5~\cite{claude}, Gemini 3.1 Pro~\cite{gemini}, Qwen3-VL-8B~\cite{qwen3vl},
five Gemma 4 variants (E2B, E4B, 12B, 26B-A4B and 31B)~\cite{gemma4}, Llama 4 Scout~\cite{llama4}
and Mistral Small 4~\cite{mistralsmall4}, without additional task-specific training.
This does not establish that their original training excluded building attributes or evaluation imagery.

The saved attribute jobs use two frozen prompts, shared by all generalists (full text in Supp.\ Sec.~S4). The street prompt requests JSON attributes from a street image,
including roof answers with a not-visible option. The roof prompt requests only roof shape and material,
adds category definitions, removes that abstention option and supplies street and outlined aerial images.
Benchmark outlines come from known building geometry. Comparing the street prompt with the roof prompt would change
the question as well as the view, so for RQ3 we also ask Claude and Gemini the roof prompt with the street photo
only. The control and street-plus-aerial conditions use
identical roof-task instructions and category definitions;
only the aerial image and the opening sentences, which
describe the supplied images and what each shows, differ.

\paragraph{Inference settings.}
Claude and Gemini run with low reasoning effort, to limit cost and stay closer to the open models, whose thinking
is turned off where possible; higher effort was not tested and may change their scores~\cite{hariri2026testtime}. The open models use greedy decoding: Qwen with 4-bit weights on an
8\,GB laptop GPU, the others served with vLLM on B200 GPUs. Image sizes and compute are therefore not matched across
models (Supp.\ Sec.~S1).

\paragraph{Specialists.}
BMAT~\cite{bmat}, RoofNet~\cite{tarver2025roofnet}, BRAILS++~\cite{brails,brailspp} and the OpenFACADES
model~\cite{openfacades} run with their released weights and input formats, BMAT with 4-bit weights (Supp.\ Sec.~S1).
A specialist is scored only on what it can answer: BRAILS++ predicts three roof shapes, so it is compared on the
flat, gable and hip roofs, and its floor detector also on the UK buildings with one to three floors. We call a test
region a specialist's \emph{home} when its training data include images from that city or region and image source.
Supp.\ Tab.~S1 lists each specialist's network, training data and published score next to ours; the datasets
and classes differ, so the comparison is a plausibility check of our setup rather than a reproduction.

\paragraph{Counting.}
Claude, Gemini and Qwen receive a counting question for
windows, exterior doors and balconies.
Grounding DINO~\cite{groundingdino} and SAM 3~\cite{sam3}
use score, area and nesting filters chosen on practice
facades. The SAM 3 car rule removes a counted box when
at least half its area lies inside a detected car or truck;
this rule applies to all three object classes.
The constant baseline uses the practice medians
(Supp.\ Sec.~S1).

\paragraph{Cohorts and scores.}
A method is eligible for an attribute when it has
saved prediction records for at least 90\% of the
labeled questions. All methods in a row use the
same cases and classes, and unparseable or abstaining
answers count as wrong. Macro-F1 averages classes
with at least five reference cases, except that
UK walls use all four classes and the restricted
roof-shape comparison uses three fixed classes.
Count error is mean absolute error.

\paragraph{Uncertainty.}
We use 2,000 paired bootstrap resamples of cases
(seed zero); the 2.5th and 97.5th percentiles give
the reported intervals~\cite{efron1993}.
A method is ahead when it has the better score in
at least 97\% of paired resamples.
This is a descriptive rule, not a calibrated or
multiplicity-corrected significance test; we call
two methods tied when neither is ahead, which
does not establish equivalence.
Further scoring details are in Supp.\ Sec.~S1.
All raw answers, labels, splits and questions are
kept, so every number can be recomputed without
running the models again.

%% file: sec/5_results.tex
\section{Results}
\label{sec:results}

\input{tables/main_results}

\subsection{Generalists and specialists (RQ1)}
\label{sec:rq1}
\textbf{No model is clearly better than Claude in any of the \NUM{\MergeMainRows} conditions}: Claude has the highest or joint-highest point score in \NUM{\ResultLeadingRowsClaude} rows, and where another model scores higher (Gemini in
\NUM{\ResultLeadingRowsGemini}, Gemma 26B-A4B in \NUM{\ResultLeadingRowsGemmaMoe}, BMAT in
\NUM{\ResultLeadingRowsBmat}), the paired difference is not clear. The rows overlap (two floor and two roof-shape
subsets, two year groupings), so they are not independent tests. Mean absolute floor error orders the models
much as macro-F1 does (Supp.\ Tab.~S5): Claude misses by \NUM{0.07} and \NUM{0.29} floors on average in the UK and OpenFACADES,
against \NUM{1.28} and \NUM{1.60} for always answering the most common class.

\begin{figure*}[t]
  \centering
  \includegraphics[width=\linewidth]{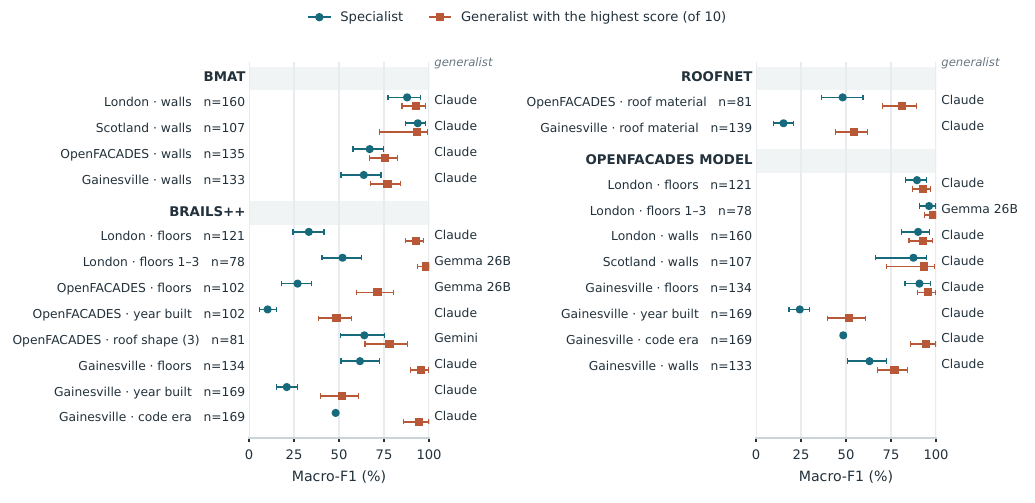}
  \caption{All \NUM{\MergeSpecialistRows} specialist comparisons: each specialist (circles) against the generalist
  with the highest macro-F1 of the ten on the same photos and answer groups (squares, named on the right), with 95\%
  bootstrap ranges and the number of photos $n$. The rows overlap and are not independent tests: London floors 1--3
  is a subset of London floors, roof shape (3) covers only flat, gable and hip roofs, and the code era (before or after
  2002) regroups the same year-built answers. Home regions (\cref{sec:methods}): London for BMAT, Florida for BRAILS++
  and RoofNet; the OpenFACADES model is away everywhere it is scored.}
  \label{fig:homeaway}
\end{figure*}

The specialists keep up only in a few settings (\cref{fig:homeaway}). Of their \NUM{\MergeSpecialistRows}
comparisons with the best generalist, \NUM{\MergeSpecialistBehind} are clearly behind. The exceptions are UK
walls, floors, and BRAILS++ on the three roof shapes it was built for. BMAT is tied with the best on Street View walls
in London, one of its training cities (\NUM{\ResultLondonWallBmat}\% vs \NUM{\ResultLondonWallClaude}\%), and in
Scotland (\NUM{\ResultScotlandWallBmat}\% vs \NUM{\ResultScotlandWallClaude}\%; this row depends on four glass
photos). It is clearly behind on the Mapillary photos of the OpenFACADES cities, which include several of its
training cities, and in Florida (\NUM{\ResultFloridaWallBmat}\% vs \NUM{\ResultFloridaWallClaude}\%). This suggests that geographic overlap alone does not guarantee similar performance; differences in image sources and reference labels may also contribute. The OpenFACADES model is tied for floors in the UK and Florida
(\NUM{\ResultUkFloorsOfmodel}\% and \NUM{\ResultFloridaFloorsOfmodel}\%) but clearly behind on Florida walls and year
built. BRAILS++, trained on US imagery, is clearly behind in Florida (\NUM{\ResultFloridaFloorsBrails}\% for floors,
\NUM{\ResultFloridaYearBrails}\% for year built); on the \NUM{\ResultOfThreeShapeCount} flat, gable and hip roofs
it is tied (\NUM{\ResultOfThreeShapeBrails}\% vs \NUM{\ResultOfThreeShapeGemini}\% for Gemini). RoofNet scores
\NUM{\ResultOfRoofRoofnet}\% on roof material in the OpenFACADES cities but \NUM{\ResultFloridaRoofRoofnet}\% in
Florida, below always answering ``asphalt shingle''. Near home, BMAT and BRAILS++ reach scores in the range their
authors report (Supp.\ Tab.~S1), which makes a broken setup less likely for them; RoofNet stays far below its
published North American accuracy in Florida.

The open models have different strengths. Gemma 26B-A4B has the highest floor point scores in OpenFACADES
(\NUM{\ResultOfFloorsGemmaMoe}\%) and the UK one-to-three-floor subset
(\NUM{\ResultUkLowFloorsGemmaMoe}\%). Its paired comparison with Claude does not establish a difference in either
condition. The 4-bit Qwen-8B scores \NUM{\ResultUkFloorsQwen}\% for UK floors and
\NUM{\ResultFloridaFloorsQwen}\% for Florida floors, but only \NUM{\ResultFloridaYearQwen}\% for Florida
construction period. Within the evaluated Gemma 4 variants, larger models do not consistently achieve higher scores across attributes
(Supp.\ Fig.~S1).

Predicting construction period across multiple decade
categories remains challenging. The highest macro-F1
is \NUM{\ResultOfYearClaude}\% in OpenFACADES and
\NUM{\ResultFloridaYearClaude}\% in Florida. Regrouping the same Florida year predictions as before 2002 versus
2002 or later gives Claude \NUM{\ResultFloridaEraClaude}\% (95\% interval
\NUM{\ResultFloridaEraLo}--\NUM{\ResultFloridaEraHi}\%). Only \NUM{\ResultFloridaNewEraCount} of the
\NUM{\ResultFloridaEraCount} retained houses belong to the later class. The binary task is easier and strongly
imbalanced; it does not establish accurate construction years or compliance with a building code.

\subsection{Label validity (RQ2)}
\label{sec:rq2}
The permit/date screen retains \NUM{139} of \NUM{154} roof-material references for the Google images and
\NUM{74} of \NUM{154} for Mapillary (\cref{fig:labels}b). For floors, construction year and walls, the corresponding
date-only retention shares are about \NUM{97}--\NUM{98}\% for Google and \NUM{85}--\NUM{86}\% for Mapillary.
Supp.\ Tab.~S2 compares scores obtained
with the applicable screening rules against those
obtained without screening. 

Removing the LiDAR agreement requirement, while keeping the date screen, expands the Google floor cohort from \NUM{134} to \NUM{169} houses;
most added houses lack a usable LiDAR comparison rather than disagree with it.
Macro-F1 decreases by \NUM{\ResultFloorScreenDropClaude} points for Claude, \NUM{\ResultFloorScreenDropGemini} for Gemini and
\NUM{\ResultFloorScreenDropQwen} for Qwen.
This LiDAR-only analysis differs from Supp.\ Tab.~S2,
which removes both screening requirements.
The reference values are unchanged, but the included houses and class proportions differ. These differences therefore measure sensitivity to reference screening and sample selection; they do not measure an improvement in label correctness against independently verified truth. Point-score ordering can also change: for Google roof material,
Gemini scores above Claude without the date screen, whereas Claude
scores slightly above Gemini with it.
This small reversal does not establish a reliable winner.

We also compare Google and Mapillary predictions on the same houses retained by both screens.
Claude's point scores are higher with the Google imagery by \NUM{\ResultPhotoFloorsGain} points for floors
(\NUM{\ResultPhotoFloorsCount} houses), \NUM{\ResultPhotoWallGain} for walls
(\NUM{\ResultPhotoWallCount}) and \NUM{\ResultPhotoRoofGain} for roof material
(\NUM{\ResultPhotoRoofCount}), but only \NUM{1.2} for year built (\NUM{149}; not clear). The paired 95\% difference
intervals are [\NUM{\ResultPhotoFloorsLo}, \NUM{\ResultPhotoFloorsHi}],
[\NUM{\ResultPhotoWallLo}, \NUM{\ResultPhotoWallHi}] and
[\NUM{\ResultPhotoRoofLo}, \NUM{\ResultPhotoRoofHi}] points, respectively; the latter two include zero, but Google
is ahead in \NUM{97}\% of paired resamples for both, just meeting our criterion.
Image provider, viewpoint, crop and capture time change together, and the roof comparison also changes the aerial
source. These results do not isolate image age.

\subsection{Street and aerial views (RQ3)}
\label{sec:rq3}
The street question and the roof question differ in wording as well as in the photos, so we separate the two effects
with the control question (\cref{tab:control}). On the \NUM{\CtrlOfRoofClaudeN} roofs with a material label in the
OpenFACADES cities, Claude scores \NUM{\CtrlOfRoofClaudeStreet}\% with the street question,
\NUM{\CtrlOfRoofClaudeControl}\% with the roof question and the street photo only, and \NUM{\CtrlOfRoofClaudeBoth}\%
when the aerial photo is added. The aerial photo alone adds \NUM{\CtrlOfRoofClaudeAerialGain} points (median paired difference; 95\% range
\NUM{\CtrlOfRoofClaudeAerialLo} to \NUM{\CtrlOfRoofClaudeAerialHi}); the better question adds
\NUM{\CtrlOfRoofClaudeQuestionGain} (\NUM{\CtrlOfRoofClaudeQuestionLo} to \NUM{\CtrlOfRoofClaudeQuestionHi}).
For Gemini, adding the aerial photo improves macro-F1 by
\NUM{\CtrlOfRoofGeminiAerialGain} points
(95\% interval \NUM{\CtrlOfRoofGeminiAerialLo} to
\NUM{\CtrlOfRoofGeminiAerialHi}).
Although the interval includes zero, the improvement meets
our descriptive 97\% paired-bootstrap criterion
(\cref{tab:control}).
The roof-specific question adds
\NUM{\CtrlOfRoofGeminiQuestionGain} points.
For roof shape, the question helps (\NUM{\CtrlOfShapeClaudeQuestionGain} points, \NUM{\CtrlOfShapeClaudeQuestionLo}
to \NUM{\CtrlOfShapeClaudeQuestionHi}) but the aerial photo shows no clear effect (\NUM{\CtrlOfShapeClaudeAerialGain}, from
\NUM{\CtrlOfShapeClaudeAerialLo} to \NUM{\CtrlOfShapeClaudeAerialHi}). For the \NUM{\CtrlFlRoofClaudeN} Florida
houses, the aerial photo adds \NUM{\CtrlFlRoofClaudeAerialGain} points to roof material, which is not clear
(\NUM{\CtrlFlRoofClaudeAerialLo} to \NUM{\CtrlFlRoofClaudeAerialHi}). Several differences may explain this: the roof of a
one-story house is largely visible from the street, the Florida aerial photos are satellite images rather than
city orthophotos, and the OpenFACADES roof references are OSM tags, which are often mapped from aerial imagery. The roof labels were screened with the street-photo date, not the aerial-photo date.

\input{tables/roof_control}

\subsection{Model agreement and voting}
\label{sec:agree}
We retrospectively examine agreement between Claude and Gemini and a plurality vote over all ten generalists
(Supp.\ Tab.~S3). The vote scores below Claude alone in all \NUM{11} evaluated conditions, by
\NUM{\ResultVoteGapMin}--\NUM{\ResultVoteGapMax} macro-F1 points; this concerns this ensemble and tie rule, not
voting in general, as five votes come from related Gemma models. Claude is more often correct where the two
commercial models agree than where they disagree (Florida roof material about \NUM{85}\% vs \NUM{40}\%; OpenFACADES
roof shape \NUM{71}\% vs \NUM{16}\%). The two models disagree on \NUM{\ResultAgreeFlagMin}--\NUM{\ResultAgreeFlagMax}\%
of cases across the evaluated conditions. This suggests a candidate review signal, not a validated confidence measure: agreeing predictions can
both be wrong, and no prospective review study was run.

\subsection{Counting windows, doors and balconies (RQ1)}
\label{sec:counting}
On \NUM{114} CMP test facades (Supp.\ Tab.~S4), SAM 3 with the car rule has the lowest window and balcony mean
absolute errors, \NUM{8.14} and \NUM{1.24} (Claude \NUM{10.04} and \NUM{1.45}; Gemini \NUM{10.35} and \NUM{1.59}),
but these differences from the commercial models do not meet our criterion. For doors, Qwen (\NUM{0.56}) and SAM 3
(\NUM{0.61}) are clearly ahead of Claude (\NUM{0.91}) and Gemini (\NUM{0.98}), whose errors are no lower than always
answering one door (\NUM{0.88}). The constant baseline scores \NUM{19.44} for windows and \NUM{3.31} for balconies,
and Grounding DINO has larger errors than SAM 3 with the car rule for all three counts. On the \NUM{13} facades with more than 60 windows, every
method miscounts by at least \NUM{40} windows on average. These results concern rectified facades, not arbitrary
street photographs.

%% file: tables/main_results.tex
\begin{table*}[t]
\centering
\footnotesize
\setlength{\tabcolsep}{2pt}
\begin{tabular}{@{}llrrcccccccccccccc@{}}
\toprule
 & & & & \multicolumn{10}{c}{Generalists (zero-shot)} & \multicolumn{4}{c}{Specialists} \\
\cmidrule(lr){5-14}\cmidrule(lr){15-18}
Region & Feature & $n$ & Ref. & C & GP & Q8 & E2 & E4 & G12 & G26 & G31 & L4 & M4 & BM & RN & BR & OF \\
\midrule
UK London & floors & 121 & 9 & \textbf{93} & 87 & 81 & 67 & 76 & 92 & 87 & 80 & 72 & 44 & -- & -- & 33 & 89 \\
UK London & floors, 1--3$^\ddagger$ & 78 & 20 & 97 & 97 & 90 & 79 & 85 & 97 & \textbf{98} & 87 & 73 & 46 & -- & -- & 52 & 96 \\
UK London & wall & 160 & 18 & \textbf{93} & 88 & 86 & 82 & 79 & 90 & 82 & 83 & 84 & 67 & 88 & -- & -- & 90 \\
UK Scotland & wall & 107 & 18 & 93 & 93 & 91 & 79 & 89 & 92 & 93 & 92 & 92 & 62 & \textbf{94} & -- & -- & 87 \\
\midrule
OpenFACADES & floors & 102 & 11 & 67 & 69 & 68 & 54 & 55 & 66 & \textbf{71} & 57 & 57 & 36 & -- & -- & 27 & $\times$ \\
 & year built & 102 & 5 & \textbf{49} & 45 & 35 & 19 & 27 & 42 & 41 & 35 & 35 & 17 & -- & -- & 10 & $\times$ \\
 & wall & 135 & 8 & \textbf{76} & 75 & 64 & 65 & 70 & 70 & 70 & 72 & 66 & 29 & 67 & -- & -- & $\times$ \\
 & roof shape & 128 & 9 & 40 & \textbf{45} & 24 & 28 & 30 & 32 & 38 & 36 & 29 & 12 & -- & -- & -- & $\times$ \\
 & roof shape, 3$^\ddagger$ & 81 & 25 & 72 & \textbf{78} & 47 & 56 & 47 & 54 & 66 & 55 & 55 & 32 & -- & -- & 64 & $\times$ \\
 & roof material & 81 & 12 & \textbf{81} & 65 & 54 & 54 & 41 & 55 & 66 & 59 & 53 & 22 & -- & 48 & -- & $\times$ \\
\midrule
Florida & floors & 134 & 46 & \textbf{95} & 95 & 89 & 77 & 84 & 92 & 95 & 93 & 91 & 42 & -- & -- & 62 & 91 \\
 & year built & 169 & 12 & \textbf{52} & 37 & 29 & 19 & 21 & 36 & 34 & 35 & 36 & 15 & -- & -- & 21 & 24 \\
 & year: code era & 169 & 48 & \textbf{95} & 75 & 64 & 48 & 48 & 75 & 72 & 78 & 58 & 48 & -- & -- & 48 & 48 \\
 & wall & 133 & 16 & \textbf{77} & 67 & 68 & 67 & 59 & 62 & 70 & 75 & 59 & 24 & 64 & -- & -- & 63 \\
 & roof material & 139 & 21 & \textbf{54} & 52 & 44 & 37 & 43 & 42 & 52 & 49 & 45 & 19 & -- & 15 & -- & -- \\
\bottomrule
\end{tabular}
\caption{Macro-F1 (\%) on the test photos. Every model in a row is scored on the same photos. All ten completed generalists and four specialists are included where applicable. \textbf{Bold}: highest point score, not a claim of statistical superiority. Ref.: descriptive test-set majority reference. UK wall rows use the same four classes, including glass. --: the model does not give this feature; $\times$: not scored (trained on these photos). C: Claude; GP: Gemini; Q8: Qwen-8B (4-bit); E2, E4, G12, G26, G31: Gemma E2B, E4B, 12B, 26B-A4B, 31B; L4: Llama Scout; M4: Mistral Small; BM: BMAT; RN: RoofNet; BR: BRAILS++; OF: OpenFACADES model. Roof rows: the generalists see the street and the aerial photo, RoofNet and BRAILS++ the aerial photo. Florida: recent photos, labels with both checks; code era: the same year answers grouped as before or after 2002. $^\ddagger$Only the classes BRAILS++ was trained on: buildings with 1--3 floors, and roofs that are flat, gable or hip (BRAILS++ is left out of the six-shape row). Complete per-model intervals accompany the evaluation outputs.}
\label{tab:main}
\end{table*}

%% file: tables/roof_control.tex
\begin{table}[t]
\centering
\footnotesize
\setlength{\tabcolsep}{3pt}
\begin{tabular}{@{}llrccc@{}}
\toprule
 & & & Street & \multicolumn{2}{c}{Roof question} \\
\cmidrule(lr){5-6}
Region, feature & Model & $n$ & question & street only & + aerial \\
\midrule
OF roof material & Claude & 81 & 42 & 56 & 81$^*$ \\
 & Gemini & 81 & 32 & 52 & 65$^*$ \\
OF roof shape & Claude & 128 & 37 & 45 & 40 \\
 & Gemini & 128 & 34 & 43 & 45 \\
Florida roof material & Claude & 139 & 51 & 50 & 54 \\
 & Gemini & 139 & 48 & 44 & 52 \\
\bottomrule
\end{tabular}
\caption{Isolating the aerial photo (RQ3): macro-F1 (\%) on the same buildings for the street question (which allows ``not visible''), the roof question with the street photo only, and the same roof question with the street and aerial photos. The last two differ only in the aerial photo and the question's opening sentences. $^*$The aerial photo helps in at least 97\% of paired bootstrap resamples.}
\label{tab:control}
\end{table}

%% file: sec/6_discussion.tex
\section{Discussion and limitations}
\label{sec:discussion}

\paragraph{What to use where.}
For describing buildings outside a specialist's home region and image source, a zero-shot commercial generalist is
the safer default: no model was clearly better than Claude in any of our conditions, while the specialists kept up
only on UK walls, on floors and on narrow tasks. Where photos may not leave an organization, an open model on one
large GPU is a middle way: Gemma 4 26B-A4B~\cite{gemma4}, of which about 4B parameters are active for each token, is best or tied
in 8 of the 15 conditions (\cref{tab:main}), including Florida floors, walls and roof material, but clearly behind on
year built in Florida and on roof material in the OpenFACADES cities.
For construction period, coarser categories aligned with practical decisions, such as construction before or
after the Florida Building Code took effect in 2002~\cite{fbc2001}, may be more useful,
although only \NUM{\ResultFloridaNewEraCount} of our Florida test houses belong to
the newer category. An aerial photo improved roof material in the OpenFACADES cities but not clearly in Florida;
it may help most where roofs are hard to see from the street. For counting windows, doors and balconies, a
promptable segmenter with a few settings chosen on practice photos is best or tied and runs on a laptop GPU. Finally, disagreement between Claude and Gemini may help
identify predictions for manual review, although its
practical value remains untested.

\paragraph{Labels.}
Screening labels changes measured scores. Without the LiDAR check, floor scores drop by six to seven points, and for
the 2014--2016 photos about half of the roof-material labels fail the photo-date screen, mostly because of a later
re-roofing permit. Such screens flag possible mismatches; they are not independent ground truth, because a permit
may not mean a changed material and a LiDAR estimate rests on assumptions about story height. We recommend that benchmarks report where labels come from,
how they were screened and how old the photos are.

\paragraph{Limitations.}
The Florida test covers one city and \NUM{\DataFloridaTest} houses,
so rare classes (flat roofs and newer houses) have limited
representation, and the screened floor cohort has no three-story houses.
Although the data were split by parcel,
some practice and test houses share source imagery. Some test examples
were inspected during early tryouts, before the protocol was fixed;
scoring corrections and the exploratory agreement and voting analyses
were conducted after inspecting test results.

Region, image source and reference-label source vary together across
datasets, so comparisons between regions are descriptive. The Google--Mapillary comparison also changes provider,
viewpoint, crop and capture time, so these comparisons do not isolate
a single cause. Overlap between model training data and public UK
and OpenFACADES evaluation imagery cannot be ruled out; the Florida
answer key was not public at evaluation time.

We used the same frozen prompts for every generalist, without per-model
tuning or few-shot examples, which might help some models, and one saved
response per model-question, with model-specific inference settings and unmatched
image resolutions and compute. The specialists were evaluated using
their released weights (BMAT in 4-bit), without local fine-tuning.
Counting was evaluated on rectified facades with connected-region
references, so the findings may not directly generalize to oblique
or occluded street scenes. The Mapillary photos were selected using
a Qwen visibility check, which may influence model comparisons.
Google imagery was used for evaluation only and is not redistributed,
apart from the credited example in \cref{fig:overview}.

\section{Conclusion}
\label{sec:conclusion}

Across three region groups, no model was clearly better than Claude
in any of the building-attribute conditions we evaluated, while
released specialist models kept up mainly on UK walls, on floors
and on the narrow tasks they were built for, even though near home
BMAT and BRAILS++ reach scores like those their authors report (RQ1).

Removing the LiDAR agreement requirement while retaining the
photo-date screen decreased Florida floor macro-F1 by six to seven
points. Model rankings changed only by small, unreliable reversals.
On the same houses, Google photos scored higher than decade-old
Mapillary photos for floors, walls and roof material, but not for year
built; provider, viewpoint and photo age change together (RQ2).

With the question held fixed, an aerial photo raised roof-material
scores in the OpenFACADES cities, with no clear effect on roof shape
or Florida roof material (RQ3).

For counting, SAM 3 achieved the lowest point errors for windows
and balconies, not clearly ahead of the commercial models; for doors,
Qwen and SAM 3 were clearly ahead of Claude and Gemini.

We release the code, questions, Florida labels and a review tool at
\url{https://github.com/mobinanoorany/building-attributes}.

\section*{Acknowledgments}
Claude (Anthropic) and ChatGPT (OpenAI) were used for language editing and to help revise the text.
The authors reviewed all content and take full responsibility for the final text, citations and numerical claims.

%% file: sec/7_supplement.tex
This supplement gives the model settings, scoring details, data details, the tables and figure that did not fit
in the paper, a floor-error table and the full prompts (\cref{sec:s4}). Section, table and figure numbers start with S; numbers without S refer to the paper.

\section{Models and scoring}
\label{sec:s1}

\paragraph{Inference settings.}
Claude and Gemini run with low reasoning effort and photos shrunk to at most 1,600 pixels on the long side. Qwen runs
on an 8\,GB laptop GPU with 4-bit weights, greedy decoding and at most about 0.79 (street) or 0.46 (roof) megapixels
per photo. The other open models are served with vLLM on B200 GPUs with greedy decoding and thinking or reasoning
turned off where the model offers a switch. Image sizes and compute are therefore not matched across models; the exact
settings and questions are released with the code.

\paragraph{Specialist adapters.}
BMAT~\cite{bmat} predicts wall material using its released fine-tuned model, with NF4 weights, the vision and output
layers kept at 16-bit, processor defaults and at most 16 greedily generated tokens. Its parser uses the first word
if it belongs to the allowed vocabulary and otherwise returns other. RoofNet~\cite{tarver2025roofnet} uses its released
CLIP weights on clean aerial imagery cropped to the known outline's bounding box with a 40-pixel margin, resized to
$224\times224$ with CLIP normalization. It selects the closest of 15 material/location text phrases, including an
unknown category. BRAILS++~\cite{brails,brailspp} uses its floor and construction-year modules on street images and
its roof-shape classifier on clean aerial images. The OpenFACADES specialist~\cite{openfacades} uses the released
InternVL3-2B checkpoint, its attribute prompt and up to six 448-pixel tiles plus a whole-image thumbnail.
\Cref{tab:repro} lists each specialist's network, training data and published score next to ours.

\input{tables/reproduction}

\paragraph{Counting.}
Only Claude, Gemini and Qwen receive the counting question. It asks for windows, exterior doors and balconies,
counts multiple panes as one window and excludes street-level shop windows. This wording is not guaranteed to
match every connected-region reference.
Grounding DINO~\cite{groundingdino} and SAM 3~\cite{sam3} use cached detections with separate practice-selected score,
area and nesting filters. Duplicate suppression uses IoU 0.5; nesting uses 80\% containment.
The SAM 3 car rule removes a counted box when at least half its area lies inside a car/truck box with detection score
at least 0.5; the implementation applies this rule to all three object classes.
The constant baseline is fitted on practice medians: 29 windows, one door and two balconies.
The count parser would replace missing fields with zero; all three saved test-answer files parse completely.

\paragraph{Cohorts and scores.}
A method is eligible for an attribute when it has saved records for at least 90\% of the labeled questions in the
corresponding job manifest. The cohort is the intersection of those methods' recorded keys, before task-specific
model exclusions. A record with an unparseable or abstaining answer remains in the cohort and is scored as wrong.
All methods within a reported row use the same cases and scored class list.
Macro-F1 averages classes having at least five reference cases, except that UK walls always use all four predefined
classes and the restricted roof slice fixes its three classes.
Rows from rare reference classes remain in the evaluated cohort and can contribute false positives to the averaged
classes. Missing answers contribute false negatives for averaged true classes. A class F1 with a zero denominator
is set to zero. The categorical majority reference is the mode of the evaluated test labels and is descriptive,
rather than a practice-fitted classifier. Count error is mean absolute error against the region-count references.

\paragraph{Uncertainty and sensitivity analyses.}
We use 2,000 ordinary bootstrap resamples of cases, seed zero, with class lists fixed from the original cohort.
Paired comparisons share sampled indices. Categorical summaries select their reference by median bootstrap score;
counting summaries use the lowest observed MAE among displayed methods, excluding plain SAM 3 from reference selection.
Cases are not grouped by panorama or spatial cluster during resampling.
Reference-screening variants reuse predictions but change the retained cases; their class lists are fixed from the
primary screened cohort. The Google/Mapillary comparison uses the intersection of eligible houses.
Agreement compares Claude and Gemini's mapped answers; the ten-model plurality vote favors Claude in a tie when its
answer is among the tied leaders, otherwise selecting alphabetically. Both analyses are retrospective.

\section{Data details}
\label{sec:s2}

\paragraph{UK.}
The practice references comprise \NUM{\DataUkPracticeFloors} floor and \NUM{\DataUkPracticeWalls} wall examples.
The saved practice and test tables have no shared MD5 image identifiers; this establishes exact-byte separation,
not absence of near-duplicates or model-training overlap.

\paragraph{OpenFACADES.}
Our stored selection has one image per building and no shared building identifiers between practice and test.
The selection code uses class quotas, prefers images with more available attributes and prioritizes US cities.
The resulting samples are not population-representative or exactly balanced after class merging.
Attributes are parsed from the provided single-question records, with roof references added from the OSM lookup.
Local city metadata are unknown for \NUM{\DataOfUnknownCity} test images and mark one additional image as other.
Aerial-image availability produces \NUM{\DataOfRoofQuestions} street-plus-aerial questions, but only
\NUM{\ResultOfShapeCount} roof-shape and \NUM{\ResultOfRoofCount} roof-material examples enter the main comparisons
after reference and cohort selection.

\paragraph{Gainesville.}
The split-generation code uses seed zero and construction-period strata; the stored split is reused.
The cached LiDAR agreement flags retain their earlier year-level date screening.
The primary photo-date screen retains uncertain within-interval ordering and flags it; the stricter variant also
removes these cases and construction in the photo year. Google test cohorts are unchanged by the stricter rule. For
Mapillary, retained counts change from \NUM{131} to \NUM{127} for floors, \NUM{149} to \NUM{144} for year,
\NUM{116} to \NUM{111} for walls and \NUM{74} to \NUM{70} for roof material. These remain comparisons between nested
cohorts, not independent verification of historical labels. \Cref{tab:labels} gives the scores under each screen.

\input{tables/label_checks}

\paragraph{Class mappings.}
Floors are grouped as 1, 2, 3, 4 and 5+ in the UK and OpenFACADES, and 1, 2 and 3+ in Florida.
Construction year is grouped as before 1970, the 1970s, 1980s, 1990s, 2000s and 2010 or later.
The additional Florida split is before 2002 versus 2002 or later; it is a calendar-year proxy, not code compliance.
Wall groups are brick, stone, concrete/plaster, wood/siding, metal and glass, restricted to supported dataset labels.
UK cladding uses only brick, concrete/plaster, stone and glass.
Merging concrete with plaster and wood with siding reduces distinctions that are difficult to infer visually,
but also changes the target taxonomy. Roof shape uses flat, gable, hip, mansard, gambrel and other where supported;
county gable/hip references cannot distinguish gable from hip. Roof-material groups are asphalt shingle, metal,
tile, membrane and other.

\paragraph{Split independence and evaluation history.}
The practice/test boundary is by image hash for the saved UK tables and by building ID in our OpenFACADES
selection. Florida is split by parcel: nine Mapillary source-image IDs occur in both halves, covering 11 test crops,
and six Google panorama IDs occur in both halves, covering seven test parcels.
Thus different parcel IDs do not guarantee independent image content.
Earlier project tryouts inspected some eventual test examples. The class-consistency and date-precision corrections,
and the agreement/voting analysis, are retrospective; we do not describe this as a single untouched final test.
Model-training overlap is only partly known. In particular, the OpenFACADES specialist is excluded on its source
dataset, and a derived Florida key that was unpublished at evaluation time does not prove that the imagery or public records were unseen.

\section{Additional results}
\label{sec:s3}

\paragraph{Model agreement and voting.}
\Cref{tab:agree} gives the agreement and voting results summarized in the paper. For floors, the vote scores
\NUM{\ResultUkVote}\% versus Claude's \NUM{\ResultUkVoteClaude}\% in the UK and \NUM{\ResultOfVote}\% versus
\NUM{\ResultOfVoteClaude}\% in OpenFACADES.

\input{tables/agreement}

\paragraph{Counting.}
\Cref{tab:counting} gives the counting errors for all methods.

\input{tables/counting}

\begin{figure}[t]
  \centering
  \includegraphics[width=\linewidth]{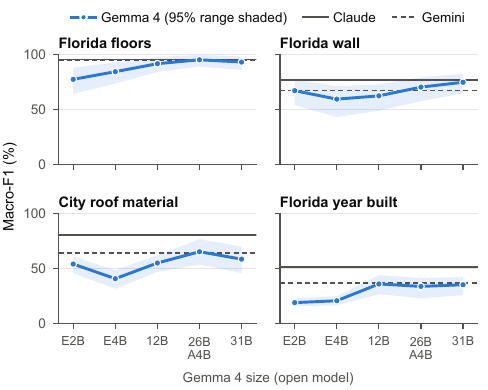}
  \caption{Gemma 4 from E2B to 31B (blue, 95\% range shaded) against Claude and Gemini on the same photos, for four
  attributes. Larger variants do not improve every attribute, and on year built and city roof material all sizes
  stay below Claude. The variants also differ in architecture, so this is not a controlled scaling study.}
  \label{fig:size}
\end{figure}

\paragraph{Gemma sizes.}
\Cref{fig:size} compares the five Gemma 4 variants with Claude and Gemini on the same photos.

\paragraph{Floor error in floors.}
Macro-F1 treats every wrong floor group alike. \Cref{tab:floormae} adds the mean absolute error in floors on the
same photos and saved answers as the floor rows of the paper's main table; no model was run again. It orders the
models much as macro-F1 does. At least 94\% of each generalist's answers are within one floor, except Mistral
Small 4 on OpenFACADES (79\%), while BRAILS++ misses by more than one floor on 33\% of the UK and 40\% of the
OpenFACADES photos.
Unlike macro-F1, MAE does not correct for class imbalance. Most Florida test houses have one story, so always
answering the most common class already reaches 0.16 floors there; the low Florida errors therefore do not show
that floors are easier to read or that the models work better in Florida.

\input{tables/floor_mae}

\section{Prompts}
\label{sec:s4}
The questions were written for this study and checked on practice photos before the test runs; earlier tryouts
also saw some eventual test examples (\cref{sec:s2}). Every generalist receives the same text, without per-model
tuning or few-shot examples. Answers are read from the last valid JSON object, with a fixed synonym list (for
example, stucco is read as plaster); missing or unknown values count as wrong. The street question also asks about
a porch or entry stairs, which is not scored.

\paragraph{Street question} (one Street View or Mapillary photo; floors, year built, wall material, and the
street-only roof answers):
\begin{lstlisting}
You are a building inspector looking at a street-view photo. Describe the main building: the one in the center of the photo.

Answer with JSON only, using exactly these keys:
{
  "floors": number of floors above ground (integer, count the rows of windows; do not count the roof),
  "year_built": approximate construction year (4-digit integer),
  "wall_material": main material of the front wall, one of "brick", "stone", "concrete", "plaster", "wood", "metal", "glass" (plaster includes stucco and painted render),
  "roof_type": one of "gable", "hip", "flat", "mansard", "gambrel", "other", "not visible",
  "roof_material": one of "asphalt shingle", "metal", "tile", "membrane", "other", "not visible" (membrane = flat roof covering such as tar paper, bitumen or gravel),
  "porch_stairs": "yes" if the main building has a porch or exterior entry stairs, otherwise "no"
}
Use "not visible" for roof_type and roof_material when the roof cannot be seen in the photo.
\end{lstlisting}

\paragraph{Roof question} (the street photo, then the aerial photo with the building outlined in yellow):
\begin{lstlisting}
The first photo is a street view of a building. The second photo is a top-down aerial photo of the same building, outlined in yellow. Use both photos to judge its roof: the aerial photo shows the roof shape and material from above, the street view shows how steep the sides are.

Roof types: "gable" = two sloping sides meeting at a ridge, with triangular ends; "hip" = slopes on all four sides; "flat" = no visible slope (often gravel, a dark membrane, solar panels or equipment on top); "mansard" = steep lower slopes on all sides with a flat or low-sloped top; "gambrel" = two different slopes on each side, barn-like; "other" = any other shape (single slope, dome, round, very mixed).
Roof materials: "asphalt shingle", "metal" (sheets or standing seams), "tile" (clay or concrete tiles, slate), "membrane" (flat roof covering such as tar paper, bitumen, gravel or concrete), "other".

Answer with JSON only:
{
  "roof_type": one of "gable", "hip", "flat", "mansard", "gambrel", "other",
  "roof_material": one of "asphalt shingle", "metal", "tile", "membrane", "other"
}
\end{lstlisting}

\paragraph{RQ3 control.} The same roof question with only the street photo; its opening sentences are replaced by
\begin{lstlisting}
The photo is a street view of a building. Use it to judge its roof.
\end{lstlisting}

\paragraph{Counting question} (CMP facades; Claude, Gemini and Qwen):
\begin{lstlisting}
You are a building inspector looking at a photo of the front of a building. On the main building in the photo, count:
- windows: every window (a window with several panes counts once; do not count shop windows at street level),
- doors: every exterior door,
- balconies: every balcony (each separate balcony counts once).

Answer with JSON only, using exactly these keys:
{"windows": integer, "doors": integer, "balconies": integer}
\end{lstlisting}

%% file: tables/reproduction.tex
\begin{table}[t]
\centering
\scriptsize
\setlength{\tabcolsep}{2pt}
\begin{tabular}{@{}p{0.19\linewidth}p{0.18\linewidth}p{0.18\linewidth}p{0.18\linewidth}p{0.20\linewidth}@{}}
\toprule
Specialist: task & Network & Trained on & Published & Ours \\
\midrule
BMAT~\cite{bmat}: wall & Qwen2.5-VL-7B, fine-tuned & Street View, 73 cities incl.\ London & F1 91 & 88 London, 94 Scotland, 64 Florida \\
RoofNet~\cite{tarver2025roofnet}: roof material & RemoteCLIP ViT-L/14, fine-tuned & satellite tiles, 101 countries & acc.\ 48; 44 N.~America & 49 OF cities, 17 Florida \\
BRAILS++~\cite{brails,brailspp}: floors & EfficientDet-D4 & US street photos & bal.\ acc.\ 67; 53 Houston~\cite{zeroshot} & 60 Florida, 57 London, 1--3 floors \\
BRAILS++: year built & ResNet-50 & US street photos & bal.\ acc.\ 18~\cite{zeroshot} & 23 Florida \\
BRAILS++: roof shape & EfficientNetV2-S & US satellite photos & acc.\ 93 (test)~\cite{brailsroof} & 72 OF cities, 3 shapes \\
OF-model~\cite{openfacades}: floors, year, wall & InternVL3-2B, fine-tuned & Mapillary, 7 cities & wall macro-F1 59 & 90 London, 63 Florida \\
\bottomrule
\end{tabular}
\caption{The specialist models: network, training data, and the score their authors report on their own data next to ours, with the stated metric (F1 or macro-F1, accuracy, or balanced accuracy). Ours uses the published metric, so it can differ from the macro-F1 in the paper's main results table. The published BRAILS++ floor and year-built scores are those reported for the original BRAILS models by Pan et al.~\cite{zeroshot}. Different datasets and class definitions prevent this comparison from reproducing or validating the published results.}
\label{tab:repro}
\end{table}

%% file: tables/label_checks.tex
\begin{table*}[t]
\centering
\footnotesize
\setlength{\tabcolsep}{3pt}
\begin{tabular}{@{}llrcccc@{}}
\toprule
Photos & Feature & Houses & Claude & Gemini & Qwen-8B$^\dagger$ & Gemma-31B \\
\midrule
recent & floors & 134/173 & 95$\to$90 & 95$\to$88 & 89$\to$83 & 93$\to$87 \\
recent & year built & 169/173 & 52$\to$51 & 37$\to$37 & 29$\to$28 & 35$\to$35 \\
recent & wall & 133/137 & 77$\to$77 & 67$\to$68 & 68$\to$68 & 75$\to$75 \\
recent & roof material & 139/154 & 54$\to$53 & 52$\to$54 & 44$\to$44 & 49$\to$49 \\
recent & roof references excluded$^*$ & 15 & 67\% & 80\% & 60\% & 67\% \\
\midrule
old & floors & 131/173 & 86$\to$85 & 86$\to$82 & 78$\to$77 & 84$\to$81 \\
old & year built & 149/173 & 57$\to$57 & 55$\to$55 & 46$\to$47 & 38$\to$40 \\
old & wall & 116/137 & 59$\to$57 & 60$\to$57 & 49$\to$48 & 47$\to$46 \\
old & roof material & 74/154 & 39$\to$42 & 40$\to$43 & 30$\to$32 & 43$\to$45 \\
old & roof references excluded$^*$ & 80 & 79\% & 79\% & 78\% & 80\% \\
\bottomrule
\end{tabular}
\caption{Label-screening sensitivity (RQ2), Florida test houses. Cells: macro-F1 (\%) with both checks $\to$ without the LiDAR and photo-date checks; Houses: with checks / without. Recent: Google photos (mostly 2026); old: Mapillary photos (mostly 2014--2016) of the same houses. The retained samples differ between conditions. $^*$Agreement with the county reference on date/permit exclusions; a permit does not prove completed work. Four generalists are shown (Gemma: Gemma 4 31B); all models are retained in the accompanying score export.
$^\dagger$4-bit weights on an 8\,GB GPU.}
\label{tab:labels}
\end{table*}

%% file: tables/agreement.tex
\begin{table*}[t]
\centering
\footnotesize
\setlength{\tabcolsep}{2.5pt}
\begin{tabular}{@{}llrrrrrr@{}}
\toprule
 & & & & \multicolumn{2}{c}{Claude right} & \multicolumn{2}{c}{Macro-F1} \\
\cmidrule(lr){5-6}\cmidrule(lr){7-8}
Region & Feature & $n$ & Flag & agree & disagree & Claude & vote \\
\midrule
UK & floors & 121 & 8\% & 94\% & 80\% & 93 & 89 \\
 & wall & 267 & 3\% & 97\% & 62\% & 95 & 93 \\
\midrule
OpenFACADES & floors & 102 & 24\% & 85\% & 46\% & 67 & 67 \\
 & year built & 102 & 45\% & 57\% & 41\% & 49 & 43 \\
 & wall & 135 & 9\% & 81\% & 42\% & 76 & 70 \\
 & roof shape & 128 & 25\% & 71\% & 16\% & 40 & 33 \\
 & roof material & 81 & 22\% & 87\% & 61\% & 81 & 61 \\
\midrule
Florida & floors & 134 & 3\% & 98\% & 50\% & 95 & 92 \\
 & year built & 169 & 29\% & 82\% & 55\% & 52 & 41 \\
 & wall & 133 & 20\% & 78\% & 67\% & 77 & 74 \\
 & roof material & 139 & 11\% & 85\% & 40\% & 54 & 47 \\
\bottomrule
\end{tabular}
\caption{Exploratory agreement and voting on test predictions. Flag: share of photos where Claude and Gemini disagree. Claude right: Claude's accuracy where the two agree and where they disagree. Macro-F1 of Claude alone and the plurality vote of all ten generalists; ties favor Claude when its answer is among the tied leaders. This is not an independently validated review policy.}
\label{tab:agree}
\end{table*}

%% file: tables/counting.tex
\begin{table}[t]
\centering
\footnotesize
\setlength{\tabcolsep}{4pt}
\begin{tabular}{@{}lccc@{}}
\toprule
Method & Windows & Doors & Balconies \\
\midrule
Lazy (median) & 19.4 & 0.88 & 3.31 \\
\multicolumn{4}{@{}l}{\emph{Detectors, settings tuned on practice photos}} \\
Grounding DINO & 10.3 & 0.84 & 2.25 \\
SAM 3 + car rule & \textbf{8.1} & \underline{0.61} & \textbf{1.24} \\
\multicolumn{4}{@{}l}{\emph{Generalists, zero-shot}} \\
Claude & \underline{10.0} & 0.91 & \underline{1.45} \\
Gemini & \underline{10.4} & 0.98 & \underline{1.59} \\
Qwen-8B$^\dagger$ & 12.7 & \textbf{0.56} & \underline{1.48} \\
\bottomrule
\end{tabular}
\caption{Counting on the 114 CMP test facades: average miscount per photo (lower is better). \textbf{Bold}: lowest point error; \underline{underlined}: the point leader has fewer errors in fewer than 97\% of paired resamples, which does not establish equivalence. Detector settings were chosen on separate practice facades. Lazy: always the median count on the separate practice set. $^\dagger$free setting on an 8\,GB GPU.}
\label{tab:counting}
\end{table}

%% file: tables/floor_mae.tex
\begin{table*}[t]
\centering
\footnotesize
\setlength{\tabcolsep}{4pt}
\begin{tabular}{@{}lcccccc@{}}
\toprule
 & \multicolumn{2}{c}{UK London ($n=121$)} & \multicolumn{2}{c}{OpenFACADES ($n=102$)} & \multicolumn{2}{c}{Florida ($n=134$)} \\
\cmidrule(lr){2-3}\cmidrule(lr){4-5}\cmidrule(lr){6-7}
Method & MAE [95\%] & $\le$1 floor & MAE [95\%] & $\le$1 floor & MAE [95\%] & $\le$1 floor \\
\midrule
Majority reference & 1.28 [1.08, 1.46] & 64\% & 1.60 [1.30, 1.90] & 50\% & 0.16 [0.10, 0.22] & 100\% \\
\midrule
Claude Sonnet 5.5   & 0.07 [0.03, 0.12] & 100\% & 0.29 [0.19, 0.42] & 97\% & 0.03 [0.01, 0.06] & 100\% \\
Gemini 3.1 Pro      & 0.12 [0.07, 0.18] & 100\% & 0.30 [0.20, 0.42] & 97\% & 0.02 [0.00, 0.05]$^\ast$ & 99\% \\
Qwen3-VL-8B (4-bit) & 0.19 [0.12, 0.27] & 99\%  & 0.34 [0.23, 0.48] & 95\% & 0.07 [0.03, 0.12] & 100\% \\
Gemma 4 E2B         & 0.37 [0.27, 0.48] & 96\%  & 0.47 [0.34, 0.61] & 95\% & 0.09 [0.04, 0.14] & 100\% \\
Gemma 4 E4B         & 0.26 [0.17, 0.35] & 98\%  & 0.39 [0.27, 0.52] & 97\% & 0.07 [0.03, 0.11] & 100\% \\
Gemma 4 12B         & 0.07 [0.03, 0.12] & 100\% & 0.35 [0.24, 0.49] & 96\% & 0.05 [0.01, 0.09] & 100\% \\
Gemma 4 26B-A4B     & 0.12 [0.06, 0.17] & 100\% & 0.29 [0.18, 0.42] & 97\% & 0.03 [0.01, 0.06] & 100\% \\
Gemma 4 31B         & 0.23 [0.15, 0.32] & 98\%  & 0.40 [0.28, 0.53] & 97\% & 0.05 [0.01, 0.09] & 100\% \\
Llama 4 Scout       & 0.33 [0.25, 0.43] & 98\%  & 0.43 [0.30, 0.57] & 94\% & 0.06 [0.02, 0.10] & 100\% \\
Mistral Small 4     & 0.50 [0.40, 0.60] & 95\%  & 0.77 [0.63, 0.94] & 79\% & 0.57 [0.49, 0.66] & 99\% \\
\midrule
BRAILS++            & 1.11 [0.91, 1.30] & 67\%  & 1.46 [1.22, 1.73] & 60\% & 0.19 [0.13, 0.27] & 99\% \\
OpenFACADES model   & 0.10 [0.05, 0.16] & 100\% & $\times$ & & 0.04 [0.01, 0.08] & 100\% \\
\bottomrule
\end{tabular}
\caption{Floor-count error on the same photos and saved answers as the floor rows of the paper's main results
table. MAE: mean absolute error in floors after capping at the top answer group (5+ in the UK and OpenFACADES,
3+ in Florida), with 95\% bootstrap ranges (2,000 resamples of photos, seed zero); $\le$1 floor: share of photos
answered within one floor, where a missing answer counts as wrong. Majority reference: always the most common test
label. $^\ast$Gemini answered 132 of the 134 Florida photos; its MAE uses those 132. $\times$: not scored (trained
on these photos). Most Florida test houses have one story, so a small Florida MAE is easy to reach; macro-F1
remains the main score.}
\label{tab:floormae}
\end{table*}